\documentclass[runningheads]{llncs}
\usepackage[T1]{fontenc}
\usepackage{graphicx,verbatim}
\usepackage{amsmath}
\usepackage{booktabs}
\usepackage{hyperref}
\usepackage[flushleft]{threeparttable}
\usepackage{color}

\begin{document}
\title{MAGO-SP: Detection and Correction of Water-Fat Swaps in Magnitude-Only VIBE MRI}
\titlerunning{MAGO-SP: Correction of Water-Fat Swaps in magnitude-only MRI}
%
%\begin{comment}  %% Removed for anonymized MICCAI 2025 submission
\author{Robert Graf\inst{1,2}\orcidID{0000-0001-6656-3680} \and 	
Hendrik Möller\inst{1,2} 	\and  							
Sophie Starck\inst{2} 	\and  							
Matan Atad\inst{1,2} 	\and      							
Philipp Braun\inst{3} 	\and  							
Jonathan Stelter\inst{3} 	\and 						
Annette Peters\inst{5} 	\and 							
Lilian Krist\inst{6} 	\and
Stefan N. Willich \inst{6} 	\and
%Thomas Keil\inst{6} 	\and 	% Er hat andere aus die liste gesetzt. Berlin hat jetzt 3 Slots.	
Henry Völzke\inst{7} 	\and 							
Robin Bülow\inst{8} 	\and 							
Klaus Berger\inst{9} 	\and 
Tobias Pischon\inst{6,10} \and
Thoralf Niendorf\inst{11} 	\and 						
Johannes Paetzold\inst{4} 	\and 						
Dimitrios Karampinos\inst{3}\and  						
Daniel Rueckert\inst{2,4} 	\and 						
Jan Kirschke\inst{1} 								
}
\authorrunning{R. Graf et al.}
\institute{Department of Diagnostic and Interventional Neuroradiology, School of Medicine, 
TUM University Hospital, München, Germany. 
\and
Institut für KI und Informatik in der Medizin, TUM University Hospital, 
Technical University of Munich, München, Germany.
\and
Department of Diagnostic and Interventional Radiology, School of Medicine, 
TUM University Hospital, München 
\and
Department of Computing, Imperial College London, London, UK.
\and
Institut für Epidemiologie, Helmholtz Zentrum München, Neuherberg,
\and
Institut für Sozialmedizin, Epidemiologie und Gesundheitsökonomie, Charité - Universitätsmedizin Berlin, Berlin, Germany 
\and
Institut für Community Medicine, Abteilung SHIP-KEF, University Medicine Greifswald, Greifswald, Germany. 
\and
Institute for Diagnostic Radiology and Neuroradiology, University Medicine Greifswald, Greifswald, Germany. 
\and
Institut für Epidemiologie und Sozialmedizin, University Münster, Münster, Germany
\and 
Molecular Epidemiology Research Group and Biobank Technology Platform, Max Delbruck Centre for Molecular Medicine in the Helmholtz Association, Buch, Germany
\and
Berlin Ultrahigh Field Facility (B.U.F.F), Max Delbrück Center for Molecular Medicine in the Helmholtz Association, Berlin, Germany 
}
%\end{comment}
\begin{comment}  
\author{Anonymized Authors}  %% Added for anonymized MICCAI 2025 submission
\authorrunning{Anonymized Author et al.}
\institute{Anonymized Affiliations \\
    \email{email@anonymized.com}}
\end{comment}
    
\maketitle              % typeset the header of the contribution
\begin{abstract}
Volume Interpolated Breath-Hold Examination (VIBE) MRI generates images suitable for water and fat signal composition estimation. While the two-point VIBE provides water-fat-separated images, the six-point VIBE allows estimation of the effective transversal relaxation rate R2* and the proton density fat fraction (PDFF), which are imaging markers for health and disease. Ambiguity during signal reconstruction can lead to water-fat swaps. This shortcoming challenges the application of VIBE-MRI for automated PDFF analyses of large-scale clinical data and of population studies. This study develops an automated pipeline to detect and correct water-fat swaps in non-contrast-enhanced VIBE images.
Our three-step pipeline begins with training a segmentation network to classify volumes as “fat-like” or “water-like,” using synthetic water-fat swaps generated by merging fat and water volumes with Perlin noise. Next, a denoising diffusion image-to-image network predicts water volumes as signal priors for correction. Finally, we integrate this prior into a physics-constrained model to recover accurate water and fat signals. Our approach achieves a <~1\% error rate in water-fat swap detection for a 6-point VIBE. Notably, swaps disproportionately affect individuals in the Underweight and Class 3 Obesity BMI categories. Our correction algorithm ensures accurate solution selection in chemical phase MRIs, enabling reliable PDFF estimation. This forms a solid technical foundation for automated large-scale population imaging analysis.

% Authors must provide keywords and are not allowed to remove this Keyword section.
\end{abstract}

%https://edoc.ub.uni-muenchen.de/21072/7/Luetke-Daldrup_Charlotte.pdf
%https://mriquestions.com/uploads/3/4/5/7/34572113/ma-2008-journal_of_magnetic_resonance_imaging.pdf
%https://mriquestions.com/uploads/3/4/5/7/34572113/ideal-_original_.pdf
%Ursache: Phase wrappingprospektiven Algorithm

\section{Introduction}
\begin{figure}
\centering
\includegraphics[width=.9\textwidth]{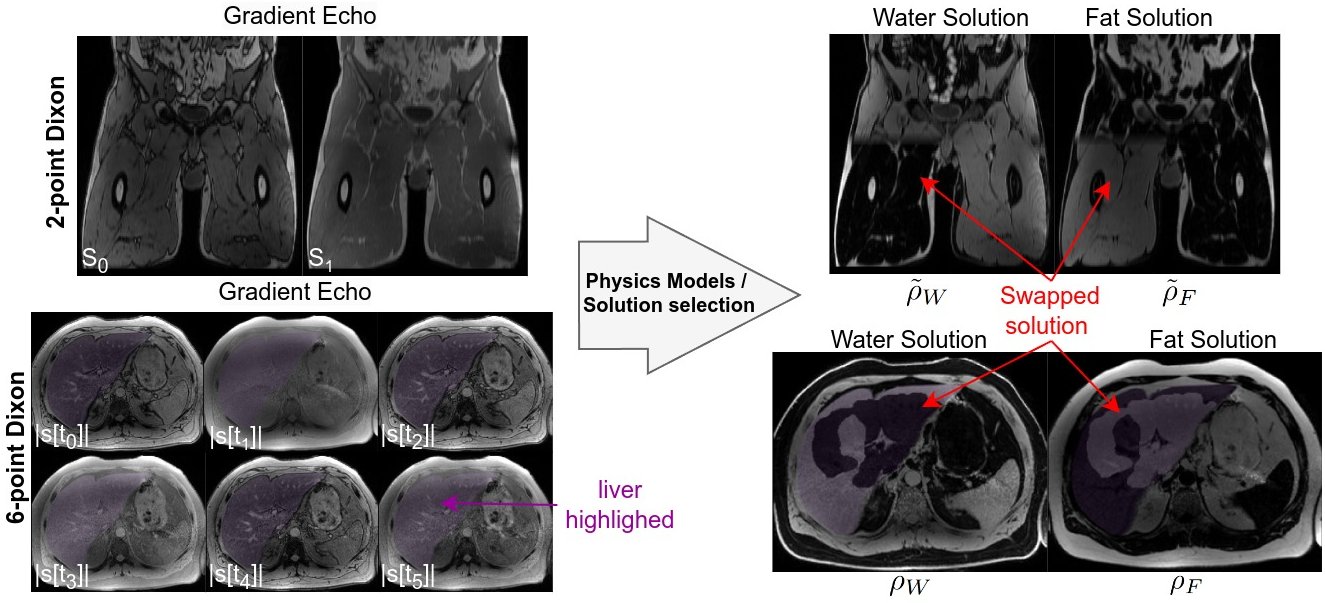}
\caption{Example of two and six-point Dixon data where the reconstruction from MRI-device vendor failed. The wrong result is selected during the solution selection due to signal ambiguity. We highlighted the liver in the 6-point Dixon in purple. } \label{fig:swapexample}
\end{figure}

Estimating fat and water fractions in soft tissue is essential, as these serve as key markers for metabolic health, body composition, and disease risk, playing a crucial role in conditions such as obesity, diabetes, and hepatic steatosis \cite{cordes2016mr,idilman2013hepatic,reeder2012proton}. The Dixon MRI technique utilizes multiple a series of MRIs acquired at different echo times (TE) to determine whether protons are bound to water or fat. Due to atomic bonding differences, fat-bound protons experience a shift in their Larmor frequency relative to water, causing them in phase and out of phase states between water and fat signals. By acquiring and reconstructing a series of MRI data as a function of the echo time TE, it is possible to estimate the effective transverse relaxation rate \( R_2^* \) and the proton densities of water (\( \rho_W \)) and fat (\( \rho_F \)) \cite{kuhn2012effect}. The general signal equation is given by:
$s[t_i] = \left(\rho_W + \rho_F \cdot \sum_{p=1}^{P} \alpha_p e^{j2\pi f_p t_i}\right) \cdot e^{j2\pi \psi t_i + \phi_0} \cdot e^{-R_2^* t_i},$
where the complex echo signal at echo time \( t_i \) is denoted as \( s[t_i] \). \( f_p \) and \( \alpha_p \) represent the frequencies and relative strengths of fat's multiple spectral peaks, and \( \psi \) accounts for field inhomogeneity. Accurate reconstruction of this signal requires parameter estimation using a fat peak model measured \textit{in vivo} \cite{fatmodel,ren2008composition,zsombor2024comparison}. Incorporating phase information helps resolve local minima and typically outperforms magnitude-only methods \cite{hernando2010chemical,reeder2005iterative}. However, clinical MRI systems commonly store and export only magnitude data, leading to the loss of phase information and necessitating magnitude-only methods. When only magnitude signals \( |s[t_i]| \) are available, the model simplifies to:
$
\left| s\left [ t_i \right ] \right| = \left|\rho_W + \rho_F \cdot \sum_{p=1}^{P} \alpha_p e^{j2\pi f_p t_i} \right| \cdot e^{-R^*_2 t_i}.
$
In the commonly used 2-point Dixon method, \( R_2^* \) effects are ignored, and a single fat peak is assumed. This further simplifies the equations to:
$
S_0 = \left|\tilde{\rho}_W - \tilde{\rho}_F\right|, \quad S_1 = \left|\tilde{\rho}_W + \tilde{\rho}_F\right|,
$
where \( \tilde{\rho}_W \) and \( \tilde{\rho}_F \) are estimates of the water and fat signals, respectively.

\begin{figure}
\centering
\includegraphics[width=.9\textwidth]{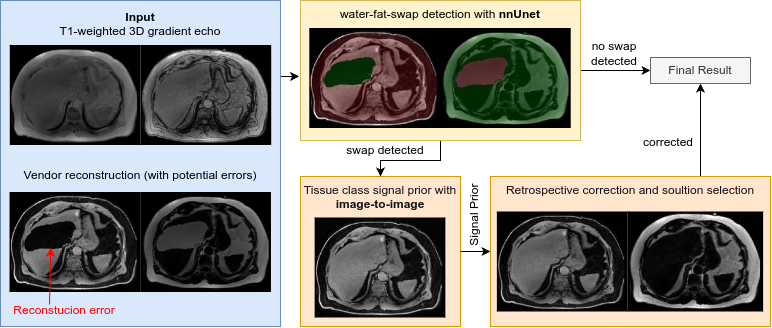}
\caption{Our proposed pipeline. We detect water-fat swaps by segmenting them into the water and fat volume. If a swap is detected, we generate a signal prior from the raw data and then use the physically constrained method to reconstruct the water and fat images. Our proposed steps are agnostic towards the number of gradient echos and the type of the applied physics model.} \label{fig:pipeline}
\end{figure}
\subsubsection{Water-fat swaps} occur when water and fat signals are incorrectly assigned due to the off-resonance effects caused by non-convex estimation of magnetic field inhomogeneities and potential additional phase errors (see Figure \ref{fig:swapexample}). In two-point VIBE, the estimated water (\(\tilde{\rho}_W\)) and fat (\(\tilde{\rho}_F\)) signals can be swapped due to inherent ambiguities in the limited data. In contrast, VIBE with more than two echoes is generally more robust against water-fat swaps. When working with complex signals, magnetic field inhomogeneities can be explicitly estimated, but phase wrapping may still introduce swaps if not properly handled \cite{ma2008dixon}. Noise and modeling inaccuracies can cause localized phase-wrapping errors to propagate, affecting larger regions or even the entire volume.
When only magnitude information is available, two-point VIBE provides insufficient data to reliably reconstruct water-fat images. The MAGO (MAGnitude‐Only) method \cite{MAGO} addresses multi-echo Dixon ambiguities by selecting the global minimum with the smallest residual among two local minima. However, this approach can yield incorrect solutions for certain pixels. Smoothing the residual helps reduce single-pixel artifacts but does not fully correct larger areas or subtle structures. 
The MAGORINO (MAGnitude‐Only with Rician noise modeling) method \cite{MAGORINO} extends MAGO by incorporating Rician noise estimation, improving accuracy in voxels with low fat signal. We further enhanced MAGO and MAGORINO by refining their multipoint search method and incorporating deep learning to predict an initial solution using a signal prior. Our signal prior can also be applied to two-point VIBE images, helping to disentangle fat and water by resolving ambiguities in the simplified physics model:
$
\tilde{\rho}_W, \tilde{\rho}_F = \left|S_0 \pm S_1\right|.
$

\subsubsection{Our objective} is to solve the automatic detection and correction of water-fat swap in a physics-aware manner \cite{jacobs2024generalizable}. We propose a processing pipeline that incorporates the following observations: Organs exhibit characteristic water or fat signals, allowing a definition of expected values to resolve ambiguities. A segmentation model can classify whether an observed region belongs to the fat or water solution for a region of interest \cite{hellgren2021detection}. With these insights, we introduce a robust framework (Figure \ref{fig:pipeline}) called "MAGnitude‐Only with Signal Prior" (MAGO-SP):
(1) Train a segmentation network to classify regions as fat-like or water-like using synthetic water-fat swaps as training data. (2) Predict the water image as a "signal prior" with denoising diffusion
\cite{ho2020denoising,song2020denoising} image-to-image network. (3) Refine the reconstruction using a classical optimization approach based on magnitude-based fitting with a multi-peak fat model \cite{fatmodel,ren2008composition}. This pipeline addresses water-fat swaps by integrating segmentation, deep learning priors, and physics-based optimization. %(4) After correction, we can rerun the segmentation and detect if there are still signal errors in the volumes.

%\( f_p \) = [-3.8, -3.4, -3.1, -2.68, -2.46, -1.95, -0.5, 0.49, 0.59]

%\( \alpha_p \) = [0.08991009, 0.58341658, 0.05994006, 0.08491508, 0.05994006, 0.01498501, 0.03996004, 0.00999001, 0.05694306]
\section{Dataset}
\begin{figure}[btp]
\centering
\includegraphics[width=1\textwidth]{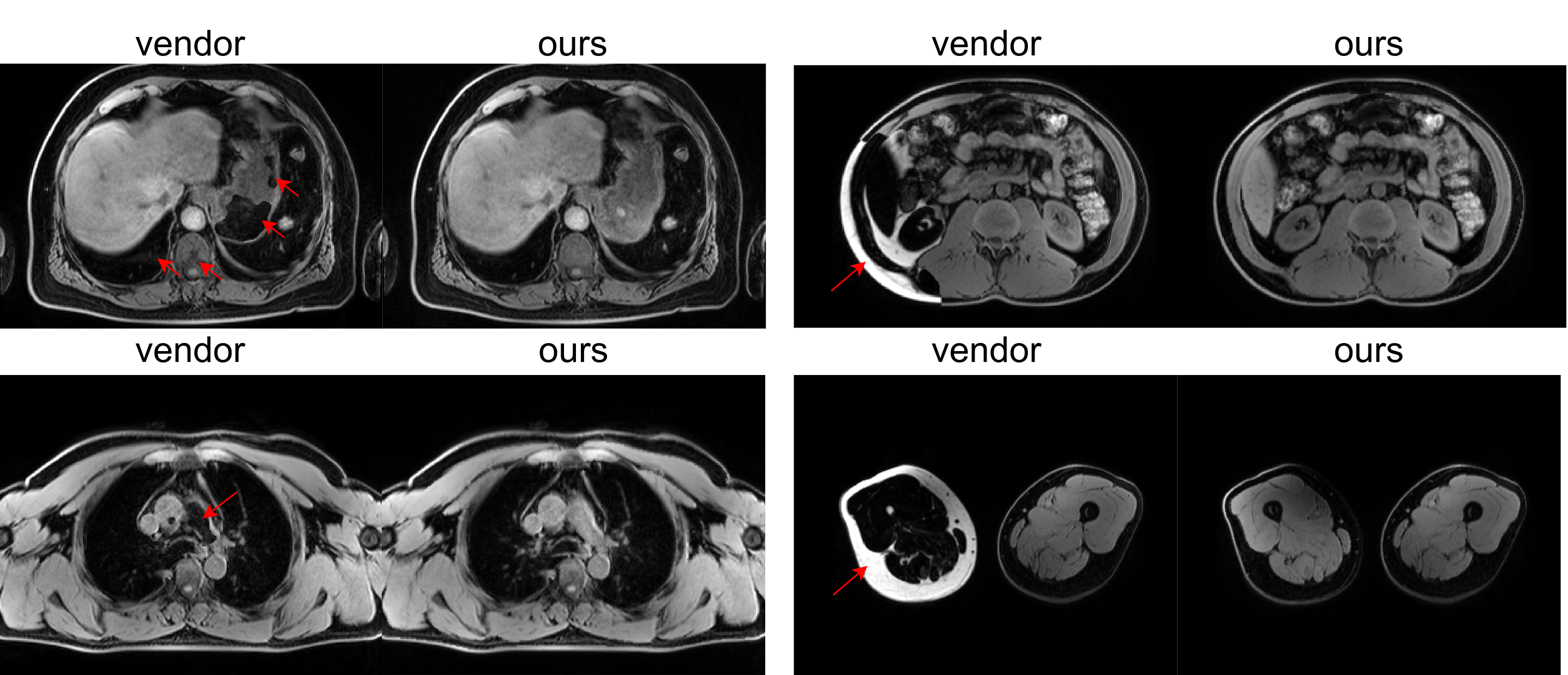}
\caption{\textit{In Vivo} water reconstruction of 2-point Dixon VIBE. Our method retrospectively corrected the swaps. Swaps are marked with red arrows. } \label{fig:vergleich2}
\end{figure}
Our experiments used 2-point- or 6-point 3D gradient echo VIBE data, with echo time $t_x = 1.23\cdot (x+1)$ ms. The MRI exam of the German National Cohort (NAKO) has 30,293 participants. This includes a 6-point VIBE MRI with the target region being the liver (axial slices, 1.64 mm in-plane, 4 mm slice thickness) and 2-point VIBE MRI with a field of view from above the knee to the neck (1.4 mm in-plane, 3 mm slice thickness). We stitched \cite{graf2024generating} the VIBE volumes together, and we counted them as one volume for our statistics. We also ran the software stack on the 52,356 2-point whole-body VIBE volumes from the UK Biobank to detect water-fat swaps (2.2 mm in-plane, 3 mm slice thickness). The UK Biobank data was acquired at 1.5T, while the NAKO data were recorded at 3.0T. We used 500 volumes from the NAKO (250 stitched 2-point and 250 6-point Dixon) for the segmentation model (20\% test set). For training the image prior, we used images without a detected swap and split them subject-wise into training (80\%; n=22649), validation (10\%; n=2734), and test split (10\%; n=2835). We use two random slices per volume during training. We classified a volume as containing water-fat swaps if at least 0.1\% of its voxels were segmented in both the water and fat reconstructions as swapped. The segmentation also highlights areas with a drop in signal integrity, leading to false PDFF values. However, this cannot be repaired because the source images are erroneous. We exclude the arms in our analysis because they are always in the region where the MR signal drops to zero, and these often have signal integrity issues. 
\section{Method}

Our method improves the magnitude-only methods by adding a swap detection through segmentation and replacing the multipoint search method with a signal prior. See Figure \ref{fig:pipeline}.

\subsubsection{Swap detection through segmentation.} 
\begin{figure}[h]
\centering
\includegraphics[width=1\textwidth]{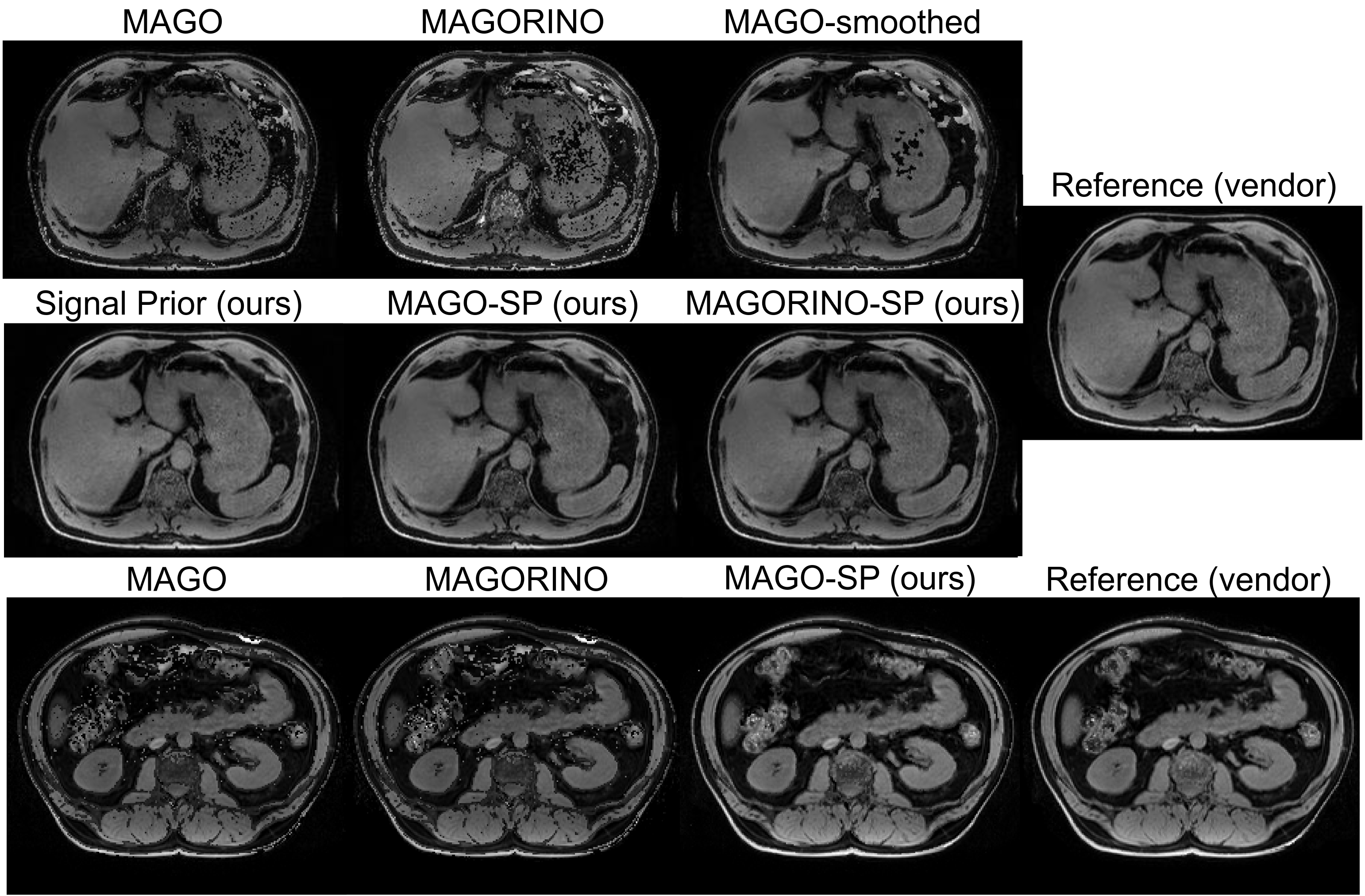}
\caption{Known good \textit{in vivo} water reconstruction of 6-point Dixon VIBE with compared reconstruction methods. False solution selection leads to black spots in the baselines. Signal prior is generated from an image-to-image network, and MAGO variants add the physical constraints.} \label{fig:vergleich}
\end{figure}
We trained a nnU-Net \cite{isensee2021nnu,isensee2024nnu} to classify whether a region belongs to a fat or water reconstruction. As auxiliary input, the first two 3D gradient-echo images ($t_0$ and $t_1$) were provided. The model predicts two output labels: water reconstruction and fat reconstruction. In cases where a patch is swapped, the segmentation should classify the affected voxels as fat in the water reconstruction and vice versa. Obtaining ground truth data with manually annotated water and fat swaps would be extremely time-consuming and impossible to do perfectly. To address this, we collected volumes without swaps and artificially introduced swaps during preprocessing \cite{hellgren2021detection}. Specifically, we generated a Perlin noise map \cite{perlin2002improving} and applied a random threshold to create a binary segmentation map, $\kappa_{\text{Perlin}}$, which served as the target segmentation. The corresponding mixed image was then synthesized as: \(
x_{\text{generated}} = x_{\text{fat}} \cdot \kappa_{\text{Perlin}} + x_{\text{water}} \cdot (1 - \kappa_{\text{Perlin}}).
\)
To generalize the model's applicability, we trained a single network using both two-point and six-point VIBE volumes. We achieved a 0.98 Dice score on our artificial test set.

\subsubsection{Signal Prior for Solutions Selection.} 
 The segmentation could already be used to swap pixels based on the predicted mask for two-point VIBE. However, errors can occur at the boundaries between swapped and non-swapped regions, leading to artifacts remaining if we only correct pixels inside the mask. For six-point VIBE, merely detecting the swap is insufficient to correct the two reconstructions. To address these challenges, we train two image-to-image generators \cite{denck2021mr}. The input consists of all 3D gradient-echo sequences, and the output is the water reconstruction. For this purpose, we employed the Palette Conditional Denoising Diffusion Network \cite{graf2023denoising,palette}. This approach generates a water image, $x_{\text{prior}}$, which serves as a signal prior. While this translation is not constrained by MRI physics (open loop), it provides a useful reference for subsequent processing. For two-point VIBE, we choose the solution from the simplified physics model that is absolute closer to  $x_{\text{prior}}$ as the water image and assign the other as the fat image. For six-point VIBE, we use the single prior, $x_{\text{prior}}$, as the initial configuration for the MAGO \cite{MAGO} and MAGORINO \cite{MAGORINO}, replacing the need for two different starting values and later disentanglement. 

After water-fat-swap correction, we can use the segmentation again and detect if signal integrity issues are left for manual review. This can detect other image errors, like signal drops. We can compute the overlap between an organ segmentation and our water-fat-swap segmentation to determine if an organ of interest was affected by a water-fat swap \cite{graf2024totalvibesegmentator}.

\section{Experiments and Results}
\begin{table}[btp]
\centering
\setlength{\tabcolsep}{2pt} % Adjust column spacing
\caption{Comparison to baselines on the test set n=600. Fraction of correctly selected solutions through our signal prior or the two-point solution method.  Structural similarity index measure (SSIM), peak signal-to-noise ratio (PSNR), and mean squared error (MSE) are between reconstruction and reference from the vendor on $\rho_W$.
}
\label{table:per}
\begin{tabular}{@{}rrrrrr@{}}
\toprule
\multicolumn{1}{c}{Method}  & \multicolumn{1}{c}{MAGO}        & \multicolumn{1}{c}{MAGO}        & \multicolumn{1}{c}{MAGORINO}    & \multicolumn{1}{c}{MAGO-SP}    & \multicolumn{1}{c}{MAGORINO} \\ 
\multicolumn{1}{l}{Metric} &             & \multicolumn{1}{c}{smoothed}    &             & \multicolumn{1}{c}{(ours) }    & \multicolumn{1}{c}{-SP(ours)}\\ \midrule
Fraction Correct $\uparrow$  & 0.825±0.02& 0.860±0.02& 0.821±0.02& \textbf{0.911±0.02}& 0.898±0.02 \\
SSIM $\rho_W$ $\uparrow   $  & 0.778±0.06& 0.815±0.04& 0.769±0.06& \textbf{0.909±0.02}& 0.885±0.02 \\
PSNR $\rho_W$ $\uparrow   $  &17.56±1.62~&18.31±1.47~&17.20±1.54~&\textbf{24.27±1.39}~&23.27±1.23~ \\
MSE  $\rho_W$ $\downarrow $  & 0.019±0.01& 0.016±0.01& 0.020±0.01& \textbf{0.004±0.00}& 0.005±0.00 \\
\midrule
\multicolumn{3}{l}{MAE  PDFF in percent:}\\
Liver $\downarrow$          & 3.41±3.05& 2.04±2.03& 3.34±3.05& 1.67±1.01& \textbf{1.63±0.89} \\
Autochthon $\downarrow$     & 6.78±1.94& 4.29±1.62& 6.59±1.91& 3.07±0.77& \textbf{2.92±0.73} \\
Vertebra $\downarrow$       &16.29±3.19&11.63±3.44&16.02±3.29&\textbf{6.30±1.22}& 7.29±1.86 \\
Kidney $\downarrow$         & 3.12±1.10& 1.87±0.59& 3.10±1.09& 1.59±0.31& \textbf{1.56±0.31} \\
\bottomrule
\end{tabular}
\end{table}
\subsubsection{Solution Selection.}

To our knowledge, 2-point Dixon magnitude disentangling does not exist in prior work. See Figure \ref{fig:vergleich} for examples. We compare different magnitude-only methods for our 6-point Dixon data. We compute the fraction of correctly assigned solutions on test set volumes without a water-fat swap in Table \ref{table:per}. A voxel is counted as correct if the absolute difference between the reconstructed water voxel and the reference water voxel is smaller than that of the reference water voxel. PDFF for air and other dark voxels can not be computed because noise dominates the signal and leads to arbitrary results. Because of this issue, we use $\rho_W$ for the visual quality metrics. We always removed voxels with a low signal for consideration and removed pixels outside the body via full body segmentation \cite{graf2024totalvibesegmentator}. Our approach is the first "magnitude-only" method that does not produce a speckle pattern, as seen in Figure \ref{fig:vergleich}. 

Our method is closer to the Vendor PDFF on known-good samples than the other methods. MAGORINO overestimates $\rho_W$ compared to the Vendor $\rho_W$. MAGORINO variants are closer to the Vendor PDFF in soft tissue, while the vertebra voxel results are inconsistent. All vertebra reconstructions are quite different and noisy, even the reference. 
We evaluate the capability of our water-fat swap detection by manually counting incorrect predictions. See Table \ref{table:userstudy}. We analyzed all 1,003 6-Point Dixon volumes identified with predicted water-fat swaps. Of these, 879 volumes were fully inverted. The segmentation network also identified additional reconstruction artifacts, including wraparounds (n=16), local signal absorption (n=1), and noise in air (n=3, counted as false). Additionally, we observed six cases where breast implants were incorrectly identified as water-fat swaps, representing false positives. In the two-point VIBE, we identified two false positives involving regions below the knee, which are usually not present in NAKO volumes. In these cases, the noise outside the body was falsely highlighted. Additionally, we randomly looked at 100 6-Point VIBE and 100 2-Point VIBE volumes and found a singular undetected water-fat-swap case. It was a walnut-sized swap close to the lung. 
%TODO ad VIBE FDR

\begin{table}[btp]
\centering
\setlength{\tabcolsep}{4pt} % Adjust column spacing
\caption{User Study: Review of Corrected Water-Fat Swap and Image Error Prediction per Full Body Volumes (Excluding Arms)}
\label{table:userstudy}
\begin{tabular}{@{}lrrrr@{}}
\toprule
Dataset & \multicolumn{2}{c}{NAKO 2-point VIBE} & \multicolumn{2}{c}{NAKO 6-point VIBE} \\ 
        \cmidrule(lr){2-3}                      
        \cmidrule(lr){4-5}
       & percent          & count/total        & percent & count/total  \\ \midrule
False Detection $\downarrow$ & 1.8  \% & (2/109)  & 0.9 \% & (9/1003)  \\
False Negative  $\downarrow$ & 0.0  \% & (0/100)  & 1.0 \%  & (1/100)    \\
\bottomrule
\end{tabular}

\end{table}

\subsubsection{Impact.}
\begin{table}[btp]
\centering
\setlength{\tabcolsep}{3pt} % Adjust column spacing
\caption{Occurrence of water-fat swaps predictions across BMI classes. Estimation of BMI-dependent biases caused by water-fat swap.}
\label{table:count}
\begin{tabular}{@{}lrrrrrr@{}}
\toprule
Dataset & \multicolumn{2}{c}{UKBB 2-point VIBE} & \multicolumn{2}{c}{NAKO 2-point VIBE} & \multicolumn{2}{c}{NAKO 6-point VIBE} \\ 
        \cmidrule(lr){2-3}                      
        \cmidrule(lr){4-5}
        \cmidrule(lr){6-7}
Images swaps      & percent & count/total & percent & count/total&percent& count/total\\ \midrule
Underweight       &39.64 \% & (155/391)   & 7.22 \% & (19/263)   & 46.77 \%& (123/263)  \\
Healthy Weight    & 8.26 \% & (1684/20376)& 2.89 \% & (364/12583)& 5.98 \% & (752/12583)\\
Overweight        & 2.57 \% & (556/21641) & 1.20 \% & (139/11581)& 0.47 \% & (55/11581) \\
Class 1 Obesity   & 1.57 \% & (117/7440)  & 0.82 \% & (36/4411)  & 0.59 \% & (26/4411)  \\
Class 2 Obesity   & 2.32 \% & (43/1857)   & 2.95 \% & (38/1286)  & 1.56 \% & (20/1286)  \\
Class 3 Obesity   & 6.61 \% & (43/651)    & 9.63 \% & (42/436)   & 6.19 \% & (27/436)   \\ 
\bottomrule
\end{tabular}
\end{table}

We analyzed the water-fat-swap rate dependency on the BMI, as we hypothesize that underweight subjects more often had swaps. In Table \ref{table:count}, we can observe that this is the case, but also towards high obesity classes. The relative error rate increases in both extremes of the BMI classes, which could lead to incorrect assessments of PDFF distributions. 

\section{Discussion and Conclusion}

This study addresses the challenge of water-fat swaps in magnitude-only VIBE MRI datasets, a problem that may introduce significant biases into PDFF measurements and downstream analyses. By proposing a fully automated pipeline, we demonstrate that combining deep learning and classical optimization enables reliable detection and correction of these swaps \textit{in vivo}.

Our approach improves upon existing magnitude multi-point Dixon methods, particularly in scenarios where current techniques like MAGO and MAGORINO struggle. They often fail to select the correct solution in \textit{in vivo} settings. MAGO is prone to underestimating PDFF in low-fat voxels due to its susceptibility to Rician noise. MAGORINO estimates the Rician noise but overestimates $\rho_W$ and $\rho_F$. Our method can replace the two-point search method with our signal prior, more reliably selecting the correct solution.\cite{zsombor2024comparison}.  Our enhancement can be incorporated into both methods. Additionally, our method can be used for 2-point Dixon data, where the water and fat signal computation is trivial, but the mathematical solution selection is no longer possible in magnitude data. Synthetic training data, generated through Perlin noise blending of known-good water and fat volumes, minimizes the need for extensive manual labeling and ensures robustness during training. 

The findings of this study reveal that water-fat swaps disproportionately affect underweight and overweight individuals, underscoring a potential source of bias in population-based studies. By automating swap detection and correction, our pipeline ensures that data from such individuals are retained, enabling more precise analyses and reducing biases across subgroups. Correcting these swaps restores the integrity of PDFF measurements, which is crucial for accurately studying conditions like hepatic steatosis. 

Our methods detect and remove water-fat errors and offer the potential to support the assessment of body fat composition in research and clinical practice. PDFF values are often used for objective determent tissue parameters and statistics in MRI.\cite{stine2021change,jung2023association} Nevertheless, there are limitations to consider. Variations between MRI vendors and reconstruction methods remain unexplored. There is currently no study that explores the differences between vendors in PDFF reconstruction. MAGO and MAGORION show that the physics and noise model can impact the reconstructed values.\cite{zsombor2024comparison}

In conclusion, this study demonstrates that a fully automated pipeline for detecting and correcting water-fat swaps \textit{in vivo} can overcome the limitations of existing methods, offering a robust and scalable solution for large datasets. %Addressing data biases supports the improvement of diagnostic imaging tools and benefits population-level analyses. 

\clearpage  % Acknowledgements, references, and appendix do not count toward the page limit (if any)
% Acknowledgments---Will not appear in the anonymized version
\bibliographystyle{splncs04}
\bibliography{bib}

\begin{thebibliography}{10}
\providecommand{\url}[1]{\texttt{#1}}
\providecommand{\urlprefix}{URL }
\providecommand{\doi}[1]{https://doi.org/#1}

\bibitem{MAGORINO}
Bray, T.J., Bainbridge, A., Lim, E., Hall-Craggs, M.A., Zhang, H.: Magorino: Magnitude-only fat fraction and r* 2 estimation with rician noise modeling. Magnetic Resonance in Medicine  \textbf{89}(3),  1173--1192 (2023)

\bibitem{cordes2016mr}
Cordes, C., Baum, T., Dieckmeyer, M., Ruschke, S., Diefenbach, M.N., Hauner, H., Kirschke, J.S., Karampinos, D.C.: Mr-based assessment of bone marrow fat in osteoporosis, diabetes, and obesity. Frontiers in endocrinology  \textbf{7}, ~74 (2016)

\bibitem{denck2021mr}
Denck, J., Guehring, J., Maier, A., Rothgang, E.: Mr-contrast-aware image-to-image translations with generative adversarial networks. International Journal of Computer Assisted Radiology and Surgery  \textbf{16},  2069--2078 (2021)

\bibitem{graf2024generating}
Graf, R., Platzek, P.S., Riedel, E.O., Kim, S.H., Lenhart, N., Ramsch{\"u}tz, C., Paprottka, K.J., Kertels, O.R., M{\"o}ller, H.K., Atad, M., et~al.: Generating synthetic high-resolution spinal stir and t1w images from t2w fse and low-resolution axial dixon. European Radiology pp. 1--11 (2024)

\bibitem{graf2024totalvibesegmentator}
Graf, R., Platzek, P.S., Riedel, E.O., Ramsch{\"u}tz, C., Starck, S., M{\"o}ller, H.K., Atad, M., V{\"o}lzke, H., B{\"u}low, R., Schmidt, C.O., et~al.: Totalvibesegmentator: Full torso segmentation for the nako and uk biobank in volumetric interpolated breath-hold examination body images. arXiv preprint arXiv:2406.00125  (2024)

\bibitem{graf2023denoising}
Graf, R., Schmitt, J., Schlaeger, S., M{\"o}ller, H.K., Sideri-Lampretsa, V., Sekuboyina, A., Krieg, S.M., Wiestler, B., Menze, B., Rueckert, D., et~al.: Denoising diffusion-based mri to ct image translation enables automated spinal segmentation. European Radiology Experimental  \textbf{7}(1), ~70 (2023)

\bibitem{fatmodel}
Hamilton, G., Schlein, A.N., Middleton, M.S., Hooker, C.A., Wolfson, T., Gamst, A.C., Loomba, R., Sirlin, C.B.: In vivo triglyceride composition of abdominal adipose tissue measured by (1) {H} {MRS} at {3T}. J Magn Reson Imaging  \textbf{45}(5),  1455--1463 (Aug 2016)

\bibitem{hellgren2021detection}
Hellgren, L., Asketun, F.: Detection of fat-water inversions in mri data with deep learning methods (2021)

\bibitem{hernando2010chemical}
Hernando, D., Liang, Z.P., Kellman, P.: Chemical shift--based water/fat separation: A comparison of signal models. Magnetic resonance in medicine  \textbf{64}(3),  811--822 (2010)

\bibitem{ho2020denoising}
Ho, J., Jain, A., Abbeel, P.: Denoising diffusion probabilistic models. Advances in neural information processing systems  \textbf{33},  6840--6851 (2020)

\bibitem{idilman2013hepatic}
Idilman, I.S., Aniktar, H., Idilman, R., Kabacam, G., Savas, B., Elhan, A., Celik, A., Bahar, K., Karcaaltincaba, M.: Hepatic steatosis: quantification by proton density fat fraction with mr imaging versus liver biopsy. Radiology  \textbf{267}(3),  767--775 (2013)

\bibitem{isensee2021nnu}
Isensee, F., Jaeger, P.F., Kohl, S.A., Petersen, J., Maier-Hein, K.H.: nnu-net: a self-configuring method for deep learning-based biomedical image segmentation. Nature methods  \textbf{18}(2),  203--211 (2021)

\bibitem{isensee2024nnu}
Isensee, F., Wald, T., Ulrich, C., Baumgartner, M., Roy, S., Maier-Hein, K., Jaeger, P.F.: nnu-net revisited: A call for rigorous validation in 3d medical image segmentation. In: International Conference on Medical Image Computing and Computer-Assisted Intervention. pp. 488--498. Springer (2024)

\bibitem{jacobs2024generalizable}
Jacobs, L., Mandija, S., Liu, H., van~den Berg, C.A., Sbrizzi, A., Maspero, M.: Generalizable synthetic mri with physics-informed convolutional networks. Medical Physics  \textbf{51}(5),  3348--3359 (2024)

\bibitem{jung2023association}
Jung, M., Rospleszcz, S., L{\"o}ffler, M.T., Walter, S.S., Maurer, E., Jungmann, P.M., Peters, A., Nattenm{\"u}ller, J., Schlett, C.L., Bamberg, F., et~al.: Association of lumbar vertebral bone marrow and paraspinal muscle fat composition with intervertebral disc degeneration: 3t quantitative mri findings from the population-based kora study. European radiology  \textbf{33}(3),  1501--1512 (2023)

\bibitem{kuhn2012effect}
K{\"u}hn, J.P., Hernando, D., Mu{\~n}oz~del Rio, A., Evert, M., Kannengiesser, S., V{\"o}lzke, H., Mensel, B., Puls, R., Hosten, N., Reeder, S.B.: Effect of multipeak spectral modeling of fat for liver iron and fat quantification: correlation of biopsy with mr imaging results. Radiology  \textbf{265}(1),  133--142 (2012)

\bibitem{ma2008dixon}
Ma, J.: Dixon techniques for water and fat imaging. Journal of Magnetic Resonance Imaging: An Official Journal of the International Society for Magnetic Resonance in Medicine  \textbf{28}(3),  543--558 (2008)

\bibitem{perlin2002improving}
Perlin, K.: Improving noise. In: Proceedings of the 29th annual conference on Computer graphics and interactive techniques. pp. 681--682 (2002)

\bibitem{reeder2012proton}
Reeder, S.B., Hu, H.H., Sirlin, C.B.: Proton density fat-fraction: a standardized mr-based biomarker of tissue fat concentration. Journal of magnetic resonance imaging: JMRI  \textbf{36}(5), ~1011 (2012)

\bibitem{reeder2005iterative}
Reeder, S.B., Pineda, A.R., Wen, Z., Shimakawa, A., Yu, H., Brittain, J.H., Gold, G.E., Beaulieu, C.H., Pelc, N.J.: Iterative decomposition of water and fat with echo asymmetry and least-squares estimation (ideal): application with fast spin-echo imaging. Magnetic Resonance in Medicine: An Official Journal of the International Society for Magnetic Resonance in Medicine  \textbf{54}(3),  636--644 (2005)

\bibitem{ren2008composition}
Ren, J., Dimitrov, I., Sherry, A.D., Malloy, C.R.: Composition of adipose tissue and marrow fat in humans by 1h nmr at 7 tesla. Journal of lipid research  \textbf{49}(9),  2055--2062 (2008)

\bibitem{palette}
Saharia, C., Chan, W., Chang, H., Lee, C., Ho, J., Salimans, T., Fleet, D., Norouzi, M.: Palette: Image-to-image diffusion models. In: ACM SIGGRAPH 2022 conference proceedings. pp. 1--10 (2022)

\bibitem{song2020denoising}
Song, J., Meng, C., Ermon, S.: Denoising diffusion implicit models. arXiv preprint arXiv:2010.02502  (2020)

\bibitem{stine2021change}
Stine, J.G., Munaganuru, N., Barnard, A., Wang, J.L., Kaulback, K., Argo, C.K., Singh, S., Fowler, K.J., Sirlin, C.B., Loomba, R.: Change in mri-pdff and histologic response in patients with nonalcoholic steatohepatitis: a systematic review and meta-analysis. Clinical Gastroenterology and Hepatology  \textbf{19}(11),  2274--2283 (2021)

\bibitem{MAGO}
Triay~Bagur, A., Hutton, C., Irving, B., Gyngell, M.L., Robson, M.D., Brady, M.: Magnitude-intrinsic water-fat ambiguity can be resolved with multipeak fat modeling and a multipoint search method. Magn Reson Med  \textbf{82}(1),  460--475 (Mar 2019)

\bibitem{zsombor2024comparison}
Zsombor, Z., Zs{\'e}ly, B., R{\'o}nasz{\'e}ki, A.D., Stollmayer, R., Budai, B.K., Palot{\'a}s, L., B{\'e}rczi, V., Kalina, I., Maurovich~Horvat, P., Kaposi, P.N.: Comparison of vendor-independent software tools for liver proton density fat fraction estimation at 1.5 t. Diagnostics  \textbf{14}(11), ~1138 (2024)

\end{thebibliography}

\appendix
%\begin{comment}  %% Removed for anonymized MICCAI 2025 submission
\section*{Acknowledgments}
Our Code and trained models are available under \href{https://github.com/robert-graf/MAGO-SP}{https://github.com/robert-graf/MAGO-SP}

The research for this article received funding from the European Research Council (ERC) under the European Union’s Horizon 2020 research and innovation program (101045128—iBack-epic—ERC2021-COG). 

This project was conducted with data (Application No. NAKO-732) from the
German National Cohort (NAKO) (www.nako.de). The NAKO is funded by the
Federal Ministry of Education and Research (BMBF) [project funding reference
numbers: 01ER1301A/B/C, 01ER1511D, 01ER1801A/B/C/D and 01ER2301A/B/C],
federal states of Germany and the Helmholtz Association, the participating
universities and the institutes of the Leibniz Association.
We thank all participants who took part in the NAKO study and the staff of
this research initiative.
%\end{comment}
\end{document}